\documentclass[11pt,a4paper]{article}
\usepackage[hyperref]{acl2019}
\usepackage{bbm}
\usepackage{times}
\usepackage{latexsym}
\usepackage{graphicx}
\usepackage{amsmath}
\usepackage{amsfonts}
\usepackage{url}
\usepackage{mathtools}
\usepackage{verbatim}
\usepackage[utf8]{inputenc}
\usepackage[shellescape]{gmp}
\usepackage{tikz}
\usepackage{natbib}
\usepackage{caption}
\usepackage{subcaption}
\usepackage{tikz-dependency}
\usepackage{wrapfig}
\usepackage{multirow}
\usepackage{pgfplots}
\usepackage{filecontents}
\usepackage[T1]{fontenc}
\usepackage{subfiles}
\usepackage{readarray}
\usepackage{xcolor}
\usepackage{import}
\aclfinalcopy 
\def\aclpaperid{2167} 

\title{Second-Order Semantic Dependency Parsing with End-to-End Neural Networks}

\author{Xinyu Wang, Jingxian Huang, Kewei Tu\thanks{\: Corresponding Author} \\
    School of Information Science and Technology,\\
    ShanghaiTech University, Shanghai, China\\
  \texttt{\{wangxy1,huangjx,tukw\}@shanghaitech.edu.cn} \\
  }

\date{}

\begin{document}
\maketitle
\begin{abstract}
Semantic dependency parsing aims to identify semantic relationships between words in a sentence that form a graph.
In this paper, we propose a second-order semantic dependency parser, which takes into consideration not only individual dependency edges but also interactions between pairs of edges. We show that second-order parsing can be approximated using mean field (MF) variational inference or loopy belief propagation (LBP). We can unfold both algorithms as recurrent layers of a neural network and therefore can train the parser in an end-to-end manner. Our experiments show that our approach achieves state-of-the-art performance.
\end{abstract}


\section{Introduction}
\label{sec:Intro}
Semantic dependency parsing \cite{oepensemeval} aims to produce graph-structured semantic dependency representations of sentences instead of tree-structured syntactic dependency parses. Existing approaches to semantic dependency parsing can be classified as graph-based approaches and transition-based approaches. In this paper, we investigate graph-based approaches which score each possible parse of a sentence by factorizing over its parts and search for the highest-scoring parse.

Previous work in graph-based syntactic dependency parsing has shown that higher-order parsing generally outperforms first-order parsing \cite{mcdonald2006online,carreras2007experiments,koo2010efficient,ma2012fourth}. While a first-order parser scores dependency edges independently, a higher-order parser takes relationships between two or more edges into consideration. However, most of the previous algorithms for higher-order syntactic dependency \emph{tree} parsing are not applicable to semantic dependency \emph{graph} parsing, and designing efficient algorithms for higher-order semantic dependency graph parsing is nontrivial. In addition, it becomes a common practice to use neural networks to compute features and scores of parse graph components, which ideally requires backpropagation of parsing errors through the higher-order parsing algorithm, adding to the difficulty of designing such an algorithm.

In this paper, we propose a novel graph-based second-order semantic dependency parser. Given an input sentence, we use a neural network to compute scores for both first and second-order parts of parse graphs and then apply either mean field variational inference or loopy belief propagation to approximately find the highest-scoring parse graph. Both algorithms are iterative inference algorithms and we show that they can be unfolded as recurrent layers of a neural network with each layer representing the computation in one iteration of the algorithms. In this way, we can construct an end-to-end neural network that takes in a sentence and outputs the approximate marginal probability of every possible dependency edge. During training, we maximize the probability of the gold parses by using standard gradient-based methods.
Our experiments show that our approach achieves state-of-the-art performance in semantic dependency parsing and outperforms our baseline with 0.3\% and 0.4\% labeled F1 score and previous state-of-the-art model with 1.3\% and 1.4\% labeled F1 score for in-domain and out-of-domain test sets respectively. Our approach shows more advantage over the baseline when there are fewer training data and when parsing longer sentences.

\begin{figure*}[t]
\centering
\includegraphics[scale=0.45]{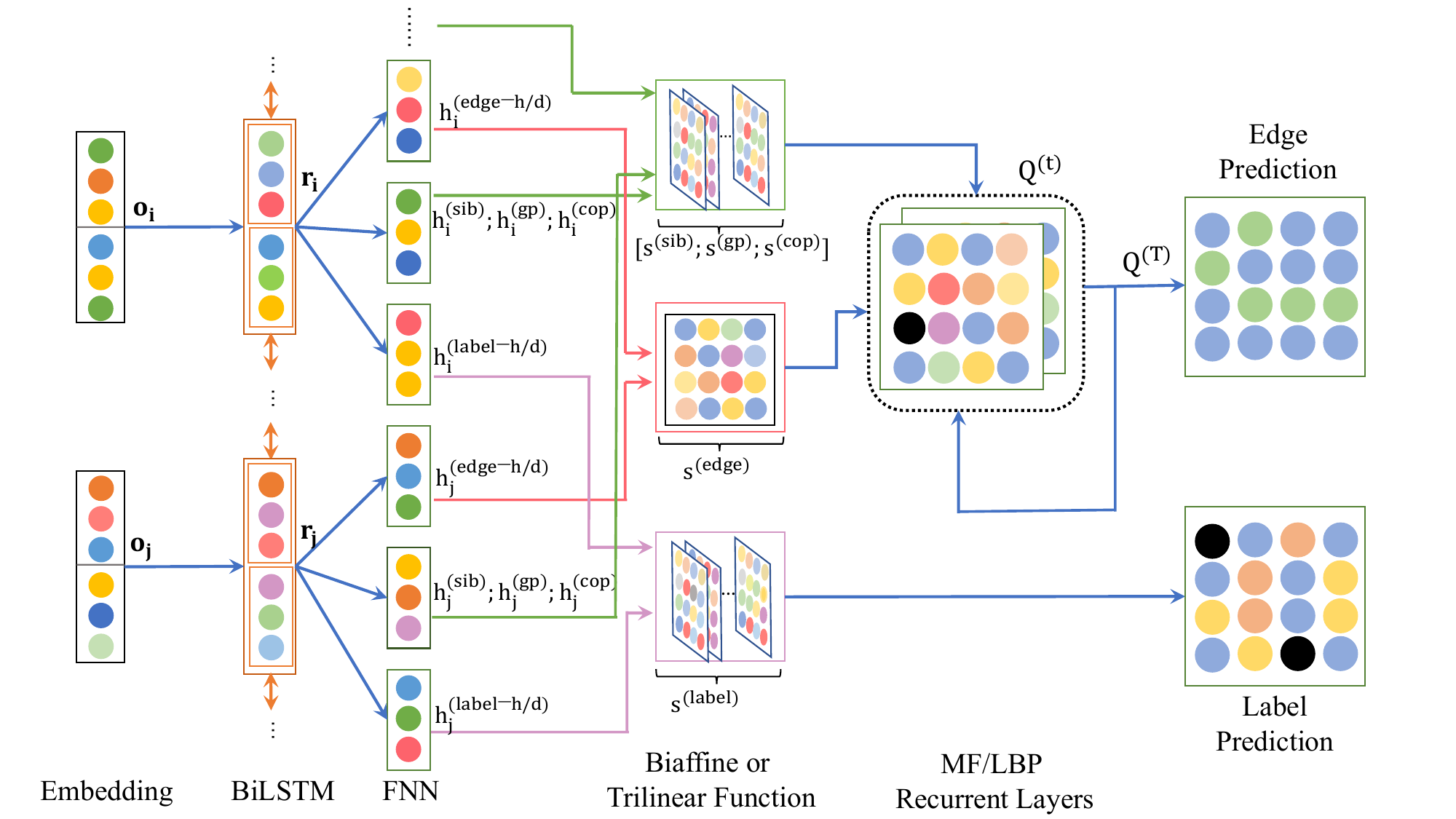}
\caption{Illustration of our model architecture.}
\label{fig:full_model}
\end{figure*}

\section{Semantic Dependency Parsing}
\label{sec:SDP}
Broad-Coverage Semantic Dependency Parsing was first defined in SemEval-2014 task 8 \cite{oepensemeval} aiming at recovering semantic dependency relationships in sentences of the WSJ corpus. It was extended in SemEval-2015 task 18 \cite{oepen2015semeval} with an additional out-of-domain dataset (the Brown corpus). A semantic dependency parse is different from a syntactic dependency parse in that the dependency edges are annotated with semantic relations (e.g., agent and patient) and form a directed acyclic graph instead of a tree. The Broad-Coverage Semantic Dependency Parsing provides three different formalisms: \textbf{DM}, \textbf{PAS} and \textbf{PSD}. Previous work has found that PAS is the easiest to learn and PSD is the most difficult as it has the largest set of labels.

\section{Approach}
\label{sec:e2e}
Our model architecture (shown in Figure \ref{fig:full_model}) follows that of \citet{dozat2018simpler}. Given an input sentence, we first compute word representations using a BiLSTM, which are then fed into two parallel modules, one for predicting the existence of every edge and the other for predicting the label of every edge. The label-prediction module makes predictions of each edge independently and hence is a first-order decoder. The edge-prediction module is what our approach differs from that of \citet{dozat2018simpler}. The module scores both first and second-order parts and then goes through multiple recurrent inference layers to predict edge existence.

\subsection{Part Scoring}
\label{sec:part score}
Given a sentence with $n$ words $[w_1,w_2,...,w_n]$, we feed a BiLSTM with their word embeddings and POS tag embeddings.
\begin{align*}
    \mathbf{o}_i&=\mathbf{e}_i^{\mathrm{(word)}}\oplus \mathbf{e}_i^{\mathrm{(postag)}}\\
    R&=\mathrm{BiLSTM}(O)
\end{align*}
where $\mathbf{o}_i$ is the concatenation ($\oplus$) of the word and POS tag embeddings of word $w_i$, $O$ represents $[\mathbf{o}_1,\dots,\mathbf{o}_n]$, and $R=[\mathbf{r}_1,\dots,\mathbf{r}_n]$ represents the output from the BiLSTM.

To score first-order parts (edges) in both the edge-prediction module and the label-prediction module, we use two single-layer feedforward networks (FNNs) to compute a \textit{head} representation and a \textit{dependent} representation for each word and then apply a biaffine function to compute the scores of edges and labels.
\begin{align}
    \mathrm{Biaff}(\mathbf{v}_1,\mathbf{v}_2)&:=\mathbf{v}_1^T \mathbf{U} \mathbf{v}_2 + \mathbf{b}\nonumber \\
    \mathbf{h}_i^{\textrm{(edge-head)}}&=\textrm{FNN}^{\textrm{(edge-head)}}(\mathbf{r}_i)\nonumber \\
    \mathbf{h}_i^{\textrm{(edge-dep)}}&=\textrm{FNN}^{\textrm{(edge-dep)}}(\mathbf{r}_i)\nonumber \\
    \mathbf{h}_i^{\textrm{(label-head)}}&=\textrm{FNN}^{\textrm{(label-head)}}(\mathbf{r}_i)\nonumber \\
    \mathbf{h}_i^{\textrm{(label-dep)}}&=\textrm{FNN}^{\textrm{(label-dep)}}(\mathbf{r}_i)\nonumber \\
    \label{eq: edge}s_{ij}^{\textrm{(edge)}}=\mathrm{Biaff}^{\textrm{(edge)}}&(\mathbf{h}_i^{\textrm{(edge-dep)}},\mathbf{h}_j^{\textrm{(edge-head)}})\\
    \label{eq: label}\mathbf{s}_{ij}^{\textrm{(label)}}=\mathrm{Biaff}^{\textrm{(label)}}&(\mathbf{h}_i^{\textrm{(label-dep)}},\mathbf{h}_j^{\textrm{(label-head)}})
\end{align}
In Eq. \ref{eq: label}, the tensor $\mathbf{U}$ in the biaffine function is $(d\times c \times d)$-dimensional and is diagonal (for any $i\neq j$, $u_{i,c,j}=0$), where $d$ is hidden size and $c$ is the number of labels. In Eq. \ref{eq: edge}, the tensor $\mathbf{U}$ in the biaffine function is $d\times 1 \times d$-dimensional.


\begin{figure}[tb]
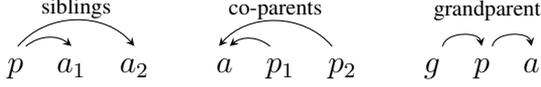

	\centering
\begin{dependency}[theme = simple]
  \begin{deptext}[column sep=0.7em, font=\large]
      $p$ \& $a_1$ \& $a_2$ \&\& $a$ \& $p_1$ \&$p_2$  \&  \&  $g$ \& $p$  \&$a$ \\
  \end{deptext}
  \depedge{9}{10}{}
  \depedge{10}{11}{}
  \depedge[arc angle=70,white,opacity = 0, text opacity = 1]{9}{11}{\large grandparent}
  \depedge[arc angle=40]{1}{2}{}
  \depedge[arc angle=50, edge start x offset=-3pt]{1}{3}{\large siblings}
  \depedge[arc angle=40,edge end x offset=2pt]{6}{5}{}
  \depedge[arc angle=50,edge end x offset=-2pt]{7}{5}{\large co-parents}
\end{dependency}
\caption{Second-order parts used in our model.}
\label{Fig:Second-order-parts}
\end{figure}

In the edge-prediction module, we further score second-order parts.
We consider three types of second-order parts: siblings (sib), co-parents (cop) and grandparents (gp) \cite{martins2014priberam}, as shown in Figure \ref{Fig:Second-order-parts}. For a specific $type$ of second-order part, we use single-layer FNNs to compute a \textit{head} representation and a \textit{dependent} representation for each word. For grandparent parts, we additionally compute a \textit{head\_dep} representation for each word.
\begin{align*}
    type&\in \{sib,cop,gp\}\\
    \mathbf{h}_i^{\text{($type$-head)}}&=\text{FNN}^{\text{($type$-head)}}(\mathbf{r}_i)\\
    \mathbf{h}_i^{\text{($type$-dep)}}&=\text{FNN}^{\text{($type$-dep)}}(\mathbf{r}_i)\\
    \mathbf{h}_i^{\text{(gp-head\_dep)}}&=\text{FNN}^{\text{(gp-head\_dep)}}(\mathbf{r}_i)
\end{align*}
We then apply a trilinear function to compute scores of second-order parts. A trilinear function is defined as follows.
\begin{align*}
\mathrm{Trilin}&(\mathbf{v}_1,\mathbf{v}_2,\mathbf{v}_3):=\mathbf{v}_3^T \mathbf{v}_1^T \mathbf{U} \mathbf{v}_2
\end{align*}
where $\mathbf{U}$ is a $(d\times d\times d)$-dimensional tensor. To reduce the computation cost, we assume that $\mathbf{U}$ has rank $d$ and can be represented as the product of three $(d\times d)$-dimensional matrices $\mathbf{U}_1$, $\mathbf{U}_2$ and $\mathbf{U}_3$. We can then compute second-order part scores as follows.
\begin{align}
        \nonumber\mathbf{g}_i:=&\mathbf{U}_i\mathbf{v}_i \qquad i\in[1,2,3]\\
        \label{eq:trilin}\mathrm{Trilin}&(\mathbf{v}_1,\mathbf{v}_2,\mathbf{v}_3):=\sum_{i=1}^{d}\mathbf{g}_{1i}\circ \mathbf{g}_{2i}\circ \mathbf{g}_{3i}\\
        \label{eq:sib}s^{(sib)}_{ij,ik} &\equiv s^{(sib)}_{ik,ij}=   \mathrm{Trilin}^{\text{(sib)}}(\mathbf{h}_i^{\text{\text{(head)}}},\mathbf{h}_j^{\text{\text{(dep)}}},\mathbf{h}_k^{\text{\text{(dep)}}})\\
        s^{(cop)}_{ij,kj} &\equiv s^{(cop)}_{kj,ij}=\label{eq:cop}\mathrm{Trilin}^{\text{(cop)}}(\mathbf{h}_i^{\text{(head)}},\mathbf{h}_j^{\text{\text{(dep)}}},\mathbf{h}_k^{\text{(head)}})\\
        s^{(gp)}_{ij,jk} &= \label{eq:gp}\mathrm{Trilin}^{\text{(gp)}}(\mathbf{h}_i^{\text{(head)}},\mathbf{h}_j^{\text{(head\_dep)}},\mathbf{h}_k^{\text{(dep)}})
\end{align}
where $\circ$ represents element-wise product. We require $j<k$ in Eq. \ref{eq:sib} and $i<k$ in Eq. \ref{eq:cop}. 

\begin{figure}[tb]
\centering
\includegraphics[scale=0.28]{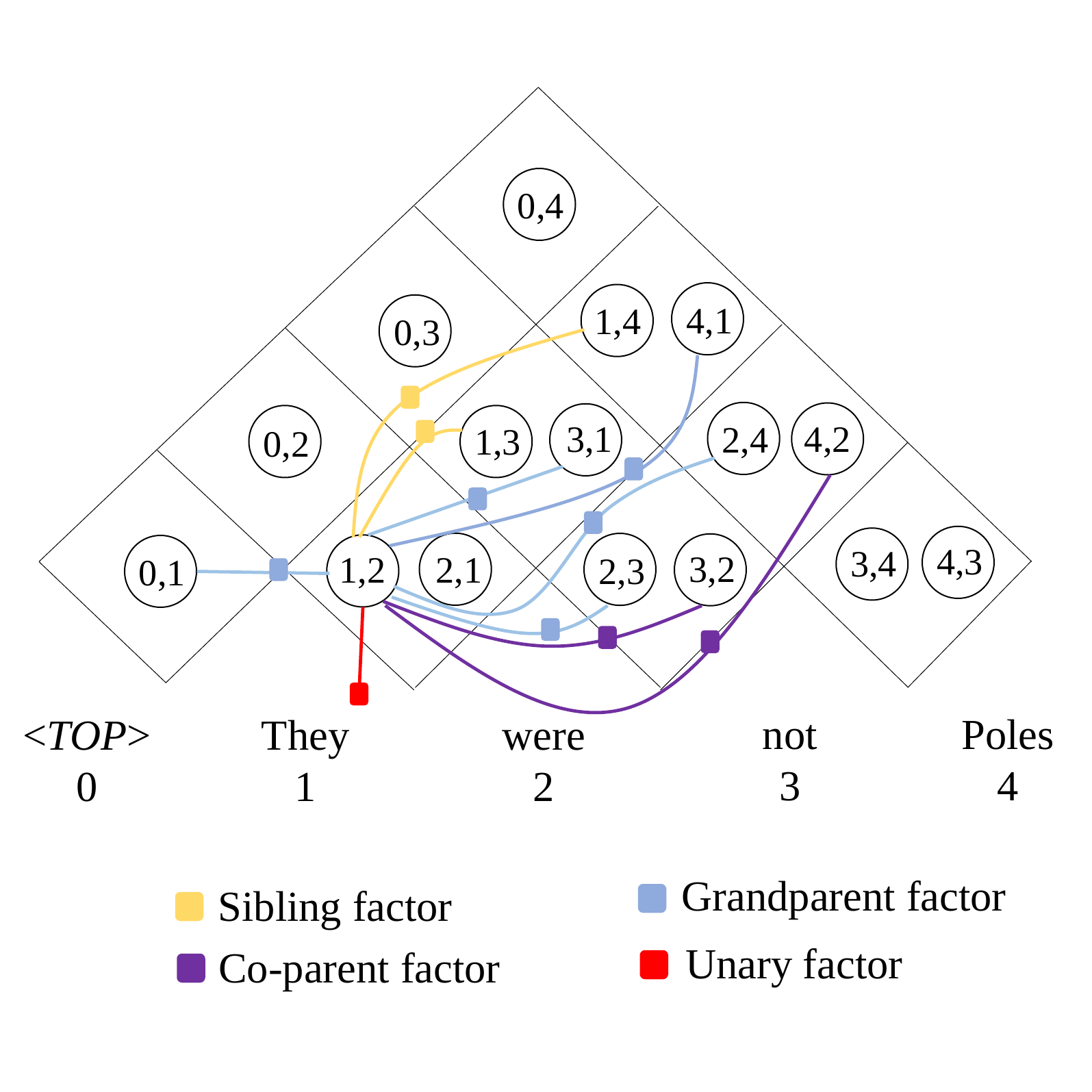}
\caption{An example of our factor graph for a sentence with four words. The node <TOP> is the top of dependency graph. The boolean variable $(i,j)$ indicates whether the directed edge $(i,j)$ exists. For simplicity, we only depict factors connected to node $(1,2)$. 
}
\label{fig:factor_graph}
\end{figure}

\subsection{Inference}
In the label-prediction module, $\mathbf{s}_{i,j}^{\text{(label)}}$ is fed into a softmax layer that outputs the probability of each label for edge $(i,j)$. In the edge-prediction module, computing the edge probabilities can be seen as doing posterior inference on a Conditional Random Field (CRF). The corresponding factor graph is shown in Figure \ref{fig:factor_graph}.
Each Boolean variable $X_{ij}$ in the CRF indicate whether the directed edge $(i,j)$ exists. We use Eq. \ref{eq: edge} to define our unary potential $\psi_u$ representing scores of an edge and Eqs. (\ref{eq:sib}-\ref{eq:gp}) to define our binary potential $\psi_p$.
We define a unary potential $\phi_u(X_{ij})$ for each variable $X_{ij}$.
\begin{align*}
    \phi_{u}(X_{ij})=&
    \begin{cases}
    \exp(s_{ij}^{\textrm{(edge)}}) &\text{$X_{ij}=1$}\\
    1 &\text{$X_{ij}=0$}
    \end{cases}
\end{align*}
For each pair of edges $(i,j)$ and $(k,l)$ that form a second-order part of a specific $type$, we define a binary potential $\phi _{p}(X_{ij}, X_{kl})$.
\begin{align*}
\phi_{p}(X_{ij},X_{kl})&=
\begin{cases}
\exp(s^{(type)}_{ij,kl}) &\text{$X_{ij}=X_{kl}=1$}\\
1 &\text{Otherwise}
\end{cases}
\end{align*}

Exact inference on this CRF is intractable. We resort to iterative approximate inference algorithms as described below, which produce the posterior distribution $Q_{ij}(X_{ij})$ of for each edge $(i,j)$. We can then predict the parse graph by including every edge $(i,j)$ such that $Q_{ij}(1) > 0.5$. The edge labels are predicted by maximizing the label probabilities computed by the label-prediction module.

\subsubsection*{Mean Field Variational Inference}
Mean field variational inference approximates a true posterior distribution with a factorized variational distribution and tries to iteratively minimize their KL divergence.
We can derive the following iterative update equations of distribution $Q_{ij}(X_{ij})$.
\begin{equation}
    \begin{aligned}
    \label{eq:msg}\mathcal{F}^{(t-1)}_{ij}=&\sum_{k\neq i,j}Q^{(t-1)}_{ik}(1)s^{(sib)}_{ij,ik}+Q^{(t-1)}_{kj}(1)s^{(cop)}_{ij,kj}\\
    &+Q^{(t-1)}_{jk}(1)s^{(gp)}_{ij,jk}+Q^{(t-1)}_{ki}(1)s^{(gp)}_{ki,ij}
    \end{aligned}
\end{equation}
\begin{equation*}
\begin{aligned}
    Q_{ij}^{(t)}(0)&\propto 1\\
    Q_{ij}^{(t)}(1)&\propto \mathrm{exp} \{s^{\textrm{(edge)}}_{ij}+ \mathcal{F}^{(t-1)}_{ij}\}\
\end{aligned}
\end{equation*}
The initial distribution $Q^{(0)}_{ij}(X_{ij})$ is set by normalizing the unary potential $\phi_u(X_{ij})$. We iteratively update the distributions for $T$ steps and then output $Q^{(T)}_{ij} (X_{ij})$, where $T$ is a hyperparameter.

\subsubsection*{Loopy Belief Propagation}
Loopy belief propagation iteratively passes messages between variables and potential functions (factors).
Because our CRF contains only unary and binary potentials, we can merge each variable-to-factor message and its subsequent factor-to-variable message into a single variable-to-variable message $M_{kl\to ij}$, representing message from edge $(k,l)$ to edge $(i,j)$. 
The update function of the messages in each iteration is:
\begin{equation*}
    Q_{ij}^{(t-1)}(X_{ij})=\phi_u (X_{ij})\prod_{ab\in\mathcal{N}_{ij}}M_{ab\to ij}^{(t-1)}(X_{ij})\\
\end{equation*}
\begin{equation*}
    \begin{aligned}
    M^{(t)}_{kl\to ij}(0)&\propto \sum_{x_{kl}} Q_{kl}^{(t-1)}(x_{kl})/M^{(t-1)}_{ij\to kl}(x_{kl})\\
    M^{(t)}_{kl\to ij}(1)&\propto Q_{kl}^{(t-1)}(0)/M^{(t-1)}_{ij\to kl}(0)\\
                        &+\exp(s^{(type)}_{ij,kl})Q_{kl}^{(t-1)}(1)/M^{(t-1)}_{ij\to kl}(1)\\
    \end{aligned}
\end{equation*}
We initialize the messages with $M^{(0)}_{kl \to ij}=1$. 
We iteratively update the messages and distributions for $T$ steps and then output normalized $Q^{(T)}_{ij} (X_{ij})$.

\subsubsection*{Inference as Recurrent Layers} \citet{zheng2015conditional} proposed that a fixed number of iterations in mean field variational inference can be seen as a recurrent neural network that is parameterized by the potential functions. We follow the idea and unfold both mean field variational inference and loopy belief propagation as recurrent neural network layers that are parameterized by part scores. 

The time complexity of our inference procedure is $O(n^3)$, which is lower than the $O(n^4)$ complexity of the exact quasi-second-order inference of \citet{cao2017quasi} and on par with the complexity of the approximate second-order inference of \citet{martins2014priberam}.

\subsection{Learning}

Given a gold parse graph $y^{\star}$ of sentence $\mathbf{w}$, the conditional distribution over possible edge $y^{\textrm{(edge)}}_{ij}$ and corresponding possible label $y^{\textrm{(label)}}_{ij}$ is given by:
\begin{align*}
&P(y^{\textrm{(edge)}}_{ij}=X_{ij}|\mathbf{w})=Q_{ij}^{(T)}(X_{ij})\\
&P(y^{\textrm{(label)}}_{ij}|\mathbf{w})=\textrm{softmax}(\mathbf{s}_{ij}^{\textrm{(label)}})
\end{align*}
We define the following cross entropy losses:
\begin{align*}
\mathcal{L}^{\textrm{(edge)}} (\theta) &= -\sum_{i,j} \log(P_\theta (y_{ij}^{\star\textrm{(edge)}}|\mathbf{w}))\\
\mathcal{L}^{\textrm{(label)}} (\theta) &= -\sum_{i,j}\mathbbm{1}(y_{ij}^{\star\textrm{(edge)}}) \log(P_\theta (y_{ij}^{\star\textrm{(label)}}|\mathbf{w}))
\end{align*}
where $\theta$ is the parameters of our model, $\mathbbm{1}(y_{ij}^{\star\textrm{(edge)}})$ denotes the indicator function and equals 1 when edge $(i,j)$ exists in the gold parse and 0 otherwise, and $i,j$ ranges over all the words in the sentence.
We optimize the weighted average of the two losses.
\begin{equation*}
    \mathcal{L}=\lambda\mathcal{L}^{(label)}+(1-\lambda)\mathcal{L}^{(edge)}
\end{equation*}
where $\lambda$ is a hyperparameter.

\section{Experiments}
\label{sec:exp}
\subsection{Hyperparameters}
We tuned the hyperparameters of our baseline model from \citet{dozat2018simpler} and our second-order model on the DM development set. We followed \citet{dozat2018simpler} using 100-dimensional pretrained GloVe embeddings \cite{pennington2014glove} and transformed them to be 125-dimensional. Words and lemmas appeared less than 7 times are replaced with a special unknown token. We use the same dataset split as in previous approaches \cite{martins2014priberam,du2015peking} with 33,964 sentences in the training set, 1,692 sentences in the development set, 1,410 sentences in the in-domain test set and 1,849 Brown Corpus sentences in the out-of-domain test set. We additionally removed sentences longer than 60 in order to speed up training, which results in 33,916 training sentences. The final hyperparameters of our baseline and second-order model are shown in Table \ref{tab:hyper_both}. Following \citet{dozat2018simpler}, we used Adam \cite{kingma2014adam} for optimizing our model, annealing the learning rate by 0.5 for every 10,000 steps, and switched the optimizer to AMSGrad \cite{reddi2018convergence} after 5,000 steps without improvement. We trained the model for 100,000 iterations with batch sizes of 6,000 tokens and terminated training early after 10,000 iterations with no improvement on the development sets.
\begin{table}[t!]
\small
\begin{center}
\begin{tabular}{lr}
\hline \hline
\textbf{Hidden Layer} & \textbf{Hidden Sizes}\\ \hline
Word/Glove/Lemma/Char & 100\\
POS & 50 \\
GloVe Linear & 125 \\
BiLSTM LSTM & 3*600 \\
Char LSTM & 1*400 \\
Unary Arc/Label & 600\\
Binary Arc & 150\\
\hline \textbf{Dropouts} & \textbf{Dropout Prob.}\\ \hline
Word/GloVe/POS/Lemma & 20\%\\
Char LSTM (FF/recur) & 33\%\\
Char Linear & 33\%\\
BiLSTM (FF/recur) & 45\%/25\%\\
Unary Arc/Label & 25\%/33\%\\
Binary Arc & 25\%\\
\hline \textbf{Optimizer \& Loss} & \textbf{Value}\\ \hline
Baseline Interpolation ($\lambda$)& 0.025\\
Second-Order Interpolation ($\lambda$)& 0.07\\
Adam $\beta_1$ & 0\\
Adam $\beta_2$ & 0.95\\
Learning rate & $1e^{-2}$\\
LR decay & 0.5\\
L2 regularization (MF/LBP) & $3e^{-9}$/$3e^{-8}$\\
\hline
\textbf{Weight Initialization} & \textbf{Mean/Stddev}\\
\hline
Unary weight (Eq. \ref{eq: edge}) & 0.0/1.0\\
Binary weight (Eq. \ref{eq:trilin}) & 0.0/0.25\\
\hline \hline
\end{tabular}
\end{center}
\caption{Hyperparameter for baseline and second-order models in our experiment. }
\label{tab:hyper_both}
\end{table}

\subsection{Main Results}

\begin{table*}[h]
\centering
\begin{tabular}{lcccccccc}
\hline \hline
 & \multicolumn{2}{c}{DM} & \multicolumn{2}{c}{PAS} & \multicolumn{2}{c}{PSD} & \multicolumn{2}{c}{Avg} \\
 & ID & OOD & ID & OOD & ID & OOD & ID & OOD \\
 \hline
Du et al. (2015)     & 89.1  & 81.8   & 91.3   & 87.2 & 75.7 & 73.3 & 85.3 & 80.8 \\
A\&M, (2015) & 88.2  & 81.8   & 90.9   & 86.9 & 76.4 & 74.8 & 85.2 & 81.2 \\
WCGL, (2018) & 90.3  & 84.9   & 91.7   & 87.6 & 78.6 & 75.9 & 86.9 & 82.8 \\
PTS17: Basic  & 89.4  & 84.5   & 92.2   & 88.3 & 77.6 & 75.3 & 86.4 & 82.7 \\
PTS17: Freda3 & 90.4  & 85.3   & 92.7   & 89.0 & 78.5 & 76.4 & 87.2 & 83.6 \\
D\&M, (2018): Basic  & 91.4  & 86.9   & 93.9   & 90.8 & 79.1 & 77.5 & 88.1 & 85.0 \\
Baseline: Basic & 92.6&	88.0&	94.1&	91.0&	80.6&	78.5&	89.1&	85.8\\
MF: Basic & \textbf{93.0}&	\textbf{88.4}&	\textbf{94.3}&	\textbf{91.5}&	80.9&	\textbf{78.9}&	\textbf{89.4}&	\textbf{86.3}\\
LBP: Basic & 92.9&	\textbf{88.4}&	\textbf{94.3}&	\textbf{91.5}&	\textbf{81.0}&	78.8&	\textbf{89.4}&	86.2\\
\hline
D\&M, (2018): +Char +Lemma & 93.7 & 88.9 & 93.9 & 90.6 & 81.0 & 79.4 & 89.5 & 86.3 \\
Baseline: +Char +Lemma & 93.7&	89.4&	94.1&	90.9&	81.0&	79.5&	89.6&	86.6 \\
MF: +Char +Lemma & \textbf{94.0}&	\textbf{89.7}&	94.1&	\textbf{91.3}&	\textbf{81.4}&	\textbf{79.6}&	\textbf{89.8}&	\textbf{86.9} \\
LBP: +Char +Lemma & 93.9&	89.5&	\textbf{94.2}&	\textbf{91.3}&	\textbf{81.4}&	79.5&	\textbf{89.8}&	86.8\\
\hline\hline
\end{tabular}
\caption{Comparison of labeled F1 scores achieved by our model and previous state-of-the-arts. The F1 scores of Baseline and our models are averaged over 5 runs. ID denotes the in-domain (WSJ) test set and OOD denotes the out-of-domain (Brown) test set. +Char and +Lemma means augmenting the token embeddings with character-level and lemma embeddings.}
\label{tab:main}
\end{table*}
We compare our model with previous state-of-the-art approaches in Table \ref{tab:main}. \citet{du2015peking} is a hybrid model. A\&M is from \citet{almeida2015lisbon}. \emph{PTS17} proposed by \cite{peng2017deep} and \emph{Basic} is single task parsing while \emph{Freda3} is a multitask parser across three formalisms. WCGL18 \cite{wang2018neural} is a neural transition-based model. D\&M \cite{dozat2018simpler} is a graph-based model and "Baseline" is the first-order model from \citet{dozat2018simpler} that was trained by ourselves. For our model, we used mean field variational inference and loopy belief propagation for 3 iterations.

In the basic setting, on average our model outperforms the best previous one by $1.3\%$ on the in-domain test set and $1.3\%$ on the out-of-domain test set. With lemma and character-based embeddings, our model leads to an average improvement of $0.3\%$ and $0.6\%$ over previous models. Our model also outperforms the baseline by $0.2\%-- 0.5\%$ on average with different settings and test sets. \citet{dozat2018simpler} found that on the PAS dataset their model cannot benefit from lemma and character-based embeddings and hence speculated that they may have approached the ceiling of the PAS F1 score. 
As shown in our experiments on the PAS dataset, our model cannot benefit from lemma and character-based embeddings either, but it obtains higher F1 scores, which suggests that the ceiling may not have been reached. 

Note that while we do not force our parser to predict a directed acyclic graph, we found that only 0.7\% of the test sentences have cycles in their parses.

\subsection{Analysis}

\subsubsection*{Small Training Data}
To evaluate the performance of our model on smaller training data, we repeated our experiments with randomly sampled 70\%, 40\% and 10\% of the training set. Table \ref{tab:samll_set} shows the F1 scores averaged over 5 runs (each time with a new randomly sampled training subset). It can be seen that the advantage of our model over the baseline increases significantly when the training data becomes smaller.
We make the following speculation to explain this observation. The BiLSTM layer in the baseline and our model is capable of capturing high-order information to some extent. However, without prior knowledge of high-order parts, it may require more training data to learn this capability than a high-order decoder. So with small training data, the baseline loses the capability of utilizing high-order information, while our model can still rely on the decoder for high-order parsing.
\begin{table*}[h]
\centering
\begin{tabular}{lcccccccc}
\hline \hline
 & \multicolumn{2}{c}{DM} & \multicolumn{2}{c}{PAS} & \multicolumn{2}{c}{PSD} & \multicolumn{2}{c}{Avg} \\
 & ID & OOD & ID & OOD & ID & OOD & ID & OOD \\
 \hline
Baseline: 70\% & 92.0&	87.0&	93.8&	90.6&	79.8&	77.7&	88.5&	85.1\\
MF: 70\% & \textbf{92.4}&	\textbf{87.5}&	93.9&	90.8&	\textbf{80.2}&	78.0&	\textbf{88.8}&	85.4\\
LBP: 70\% & 92.3&	87.4&	\textbf{94.0}&	\textbf{90.9}&	\textbf{80.2}&	\textbf{78.1}&	\textbf{88.8}&	\textbf{85.5}\\
\hline
Baseline: 40\% & 90.8&	85.5&	93.2&	89.6&	78.4&	76.4&	87.4&	83.8\\
MF: 40\% & \textbf{91.2}&	\textbf{86.0}&	93.4&	\textbf{90.0}&	\textbf{78.9}&	76.7&	87.8&	84.2\\
LBP: 40\% & \textbf{91.2}&	\textbf{86.0}&	\textbf{93.5}&	\textbf{90.0}&	\textbf{78.9}&	\textbf{76.8}&	\textbf{87.9}&	\textbf{84.3}\\
\hline
Baseline: 10\% & 86.1&	80.0&	90.8&	86.4&	73.5&	71.2&	83.4&	79.2\\
MF: 10\% & \textbf{86.9}&	\textbf{81.0}&	\textbf{91.3}&	\textbf{87.1}&	\textbf{74.5}&	72.1&	\textbf{84.2}&	\textbf{80.1}\\
LBP: 10\% & 86.8&	80.9&	\textbf{91.3}&	87.0&	\textbf{74.5}&	\textbf{72.3}&	\textbf{84.2}&	\textbf{80.1}\\
\hline\hline
\end{tabular}
\caption{Comparison of labeled F1 scores achieved by our model and our baseline on 10\%, 40\%, 70\% of the training data. We report the average F1 score over 5 runs with different randomly sampled training data.}
\label{tab:samll_set}
\end{table*}

\subsubsection*{Performance on Different Sentence Lengths} 
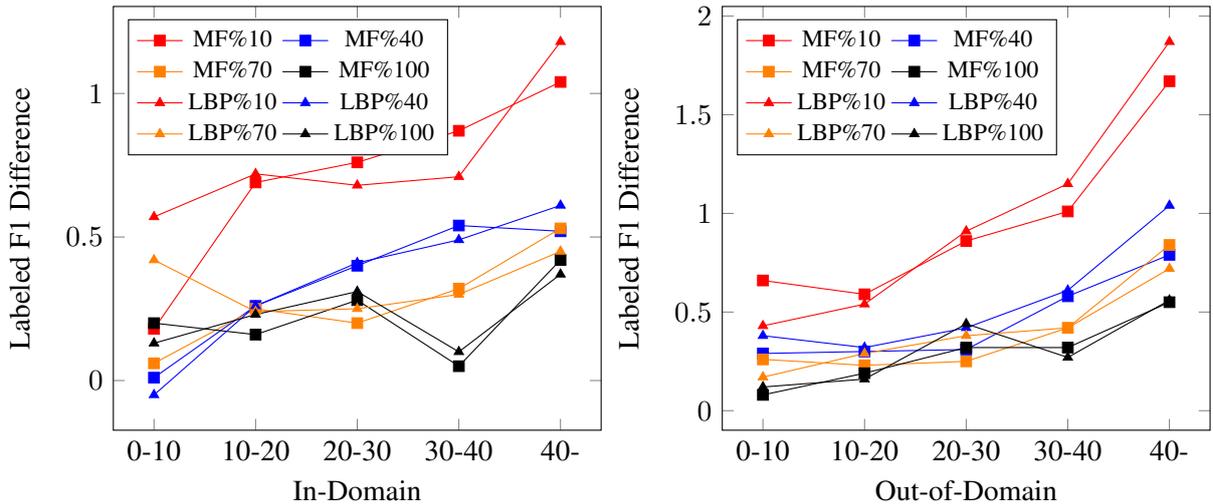
\begin{figure*}[h]
\begin{tikzpicture}
    \begin{axis}[
        width=0.5\textwidth,
        height=0.45\textwidth,
        xlabel=In-Domain,
        ylabel=Labeled F1 Difference,
        legend columns=2, 
        legend pos=north west,
        legend style={font=\small},
        xticklabels={0-10,10-20,20-30,30-40,40-},
        xtick={0,10,20,30,40},
        ytick={0,0.5,1.0,1.5,2.0,2.5,3.0}
        ]
        \addplot[red,mark=square*] table[x=length,y=MF10] {iddata.dat};
        \addplot[blue,mark=square*] table[x=length,y=MF40] {iddata.dat};
        \addplot[orange,mark=square*] table[x=length,y=MF70] {iddata.dat};
        \addplot[black,mark=square*] table[x=length,y=MF100] {iddata.dat};
        \addplot[red,mark=triangle*] table[x=length,y=LBP10] {iddata.dat};
        \addplot[blue,mark=triangle*] table[x=length,y=LBP40] {iddata.dat};
        \addplot[orange,mark=triangle*] table[x=length,y=LBP70] {iddata.dat};
        \addplot[black,mark=triangle*] table[x=length,y=LBP100] {iddata.dat};
        \legend{MF\%10,MF\%40,MF\%70,MF\%100,LBP\%10,LBP\%40,LBP\%70,LBP\%100}
    \end{axis}
\end{tikzpicture}
\begin{tikzpicture}
    \begin{axis}[
        width=0.5\textwidth,
        height=0.45\textwidth,
        xlabel=Out-of-Domain,
        ylabel=Labeled F1 Difference,
        legend columns=2, 
        legend pos=north west,
        legend style={font=\small},
        xticklabels={0-10,10-20,20-30,30-40,40-},
        xtick={0,10,20,30,40},
        ytick={0,0.5,1.0,1.5,2.0,2.5,3.0}
        ]
        \addplot[red,mark=square*] table[x=length,y=MF10] {ooddata.dat};
        \addplot[blue,mark=square*] table[x=length,y=MF40] {ooddata.dat};
        \addplot[orange,mark=square*] table[x=length,y=MF70] {ooddata.dat};
        \addplot[black,mark=square*] table[x=length,y=MF100] {ooddata.dat};
        \addplot[red,mark=triangle*] table[x=length,y=LBP10] {ooddata.dat};
        \addplot[blue,mark=triangle*] table[x=length,y=LBP40] {ooddata.dat};
        \addplot[orange,mark=triangle*] table[x=length,y=LBP70] {ooddata.dat};
        \addplot[black,mark=triangle*] table[x=length,y=LBP100] {ooddata.dat};
        \legend{MF\%10,MF\%40,MF\%70,MF\%100,LBP\%10,LBP\%40,LBP\%70,LBP\%100}
    \end{axis}
\end{tikzpicture}
\caption{Relative improvements over our baseline in different sentence length intervals with different training data sizes. We report the average F1 score improvements over all the formalisms with 5 runs for each.}
\label{fig:sentence_length}
\end{figure*}

We want to study the impact of sentence lengths on first-order parsing and our second-order parsing. We split the test sets of all the formalisms into five subsets with different sentence length ranges and evaluate our model and the baseline on them. 
Figure \ref{fig:sentence_length} shows that our model has more advantage over the baseline when sentences get longer, especially when sentences are longer than 40. One possible explanation is that BiLSTM has difficulty in capturing long-range dependencies in long sentences, which leads to lower performance on the first-order baseline; but such long-range dependencies can still be captured with second-order parsing. It can also be seen that on long sentences, our model has more advantage over the baseline for the out-of-domain test set than for the in-domain test set, which suggests that our model has better generalizability especially on long sentences.



\subsubsection*{Mean Field vs. Loopy Belief Propagation}
We compare mean field variational inference and loopy belief propagation algorithms in Table \ref{tab:model choose}. We tuned the hyperparameters of our model for each algorithm and iteration number separately. We find that in general mean field variational inference has very similar performance to loopy belief propagation. In addition, with more iterations, the performance of mean field variational inference steadily increases while the at the second iteration. 
\begin{table}[h]
\centering
\begin{tabular}{l|c|cc}
\hline \hline
 & Iteration & ID & OOD \\
 \hline
\multirow{3}{*}{MF} & 1  &  92.78 & 88.38 \\
 & 2  & 92.86 & 88.37 \\
 & 3 & \textbf{92.98} & \textbf{88.44}\\
 \hline
\multirow{3}{*}{LBP} & 1  & \textbf{92.88} & 88.29 \\
 & 2 & 92.84 & 88.17 \\
 & 3 & \textbf{92.88} & \textbf{88.36} \\
\hline \hline
\end{tabular}
\caption{Comparison of labeled F1 scores of mean field and loopy belief propagation with different iteration numbers on the DM dataset.}
\label{tab:model choose}
\end{table}

\subsubsection*{Ablation Study} 
We study how different types of second-order parts defined in Section \ref{sec:part score} affect the performance of our parser. We trained our model with each type of second-order parts without the other two types on the DM dataset using mean field variational inference and the result is shown in Table \ref{tab:ablation}. While all the three types of second-order parts can be seen to improve the parsing performance over the baseline, the sibling parts lead to the largest performance gain on both the in-domain test set and the out-of-domain test set. 
\begin{table}[t]
\centering
\begin{tabular}{l|cc}
\hline \hline
 & ID & OOD \\
 \hline
Baseline & 92.60 & 87.98 \\
+Siblings & \textbf{92.85} & \textbf{88.31} \\
+Co-parents & 92.80 & 88.23 \\
+Grandparents & 92.84 & 88.24 \\
\hline \hline
\end{tabular}
\caption{The performance comparison between three types of second-order parts on the DM dataset.}
\label{tab:ablation}
\end{table}

\subsection{Case Study} 
\begin{figure*}[h]
\centering
\import{graphs/}{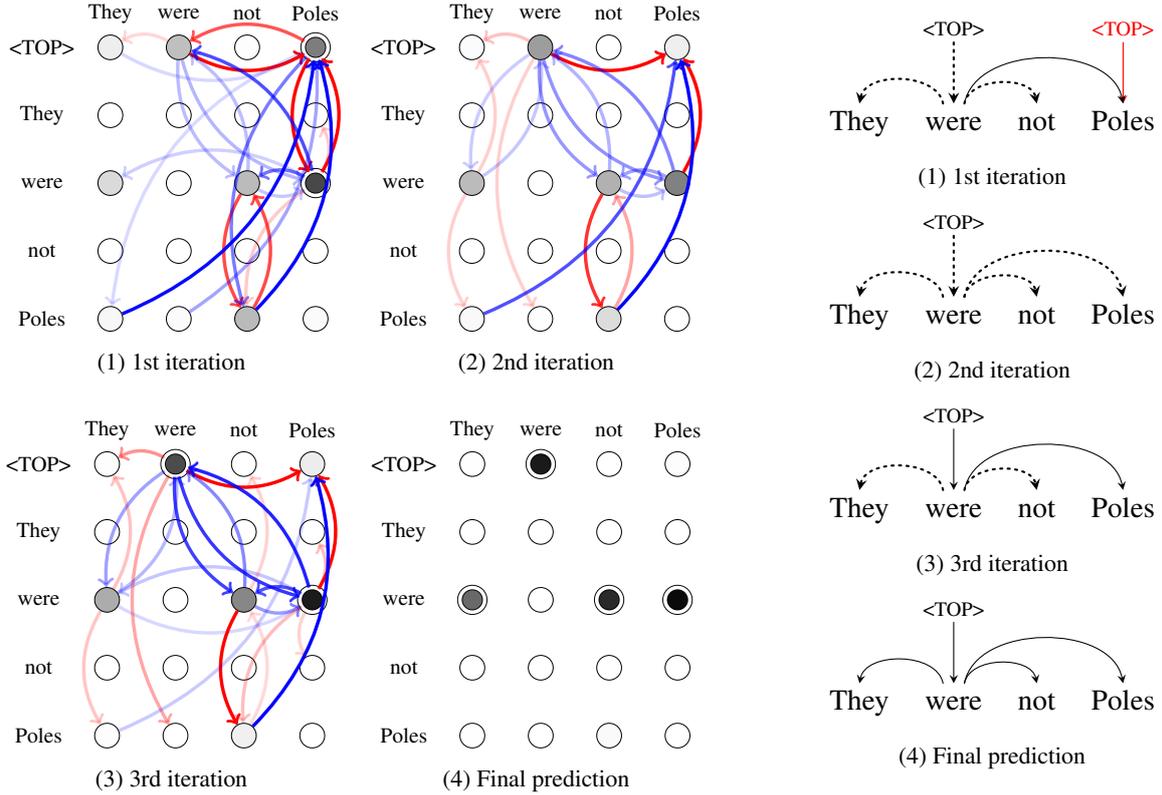}
\caption{An example of message passing (left) and the corresponding graph parses (right) in our second-order parser with mean field variational inference. We regard terms in Eq. \ref{eq:msg} as messages sent from other arcs. Blue arcs and red arcs on the left represent positive messages (which \textbf{encourage} the target edge to exist) and negative messages (which \textbf{discourage} the target edge to exist) respectively. Lightness of the arc color represents the message intensity. Blackness of each nodes represents the probability of edge existence. A Node with a double circle means the corresponding edge is predicted to exist. Messages with low intensities are omitted in the graph. Dotted arcs and red arcs on the right represent missed predictions and wrong predictions compared to the golden parse. The period in the sentence is omitted for simplicity.}
\label{fig:example}
\end{figure*}

We provide a parsing example in Figure \ref{fig:example} to show how our second-order parser with 3 iterations of mean field variational inference works. Before the first iteration, the marginal distributions of edges $Q_{ij}$ is initialized with unary potentials and thus is exactly what a first-order parser would produce. In the subsequent iterations, the distributions are updated with binary potentials taken into account. For each version of the distributions, we can extract a parse graph by collecting edges with probabilities larger than 0.5. From Figure \ref{fig:example}, we can see that erroneous edges are gradually fixed through iterations. Edge $(were,Poles)$ sends a strong negative co-parents message to edge $(\textrm{<}TOP\textrm{>},Poles)$ in the first iteration, so the latter has a lower probability in subsequent iterations. Edge $(were,Poles)$ also sends a strong positive grandparent message to edge $(\textrm{<}TOP\textrm{>},were)$ to enhance its probability, and the latter sends an increasingly positive message back to the former in subsequent iterations. In the second and third iterations, $(were,Poles)$ sends positive sibling messages and $(\textrm{<}TOP\textrm{>},were)$ sends positive grandparent messages to enhance probabilities of edges $(were,They)$ and $(were,not)$, which finally leads to the correct parse.

\begin{table}[t]
\centering
\begin{tabular}{l|ccccc}
\hline \hline
(sents/sec) & train&parse&long&short\\
 \hline
Baseline & \textbf{730}&	\textbf{904}&	\textbf{334}&	\textbf{1762}\\
MF & 472&	722&	240&	1485\\
LBP & 258&	300&	88&	1246\\
\hline \hline
\end{tabular}
\caption{Training and parsing speed (sentences/second) comparison of the baseline and our model (3 iterations for our second-order parser). \textbf{long} means the parsing speed on sentences longer than 40 and \textbf{short} means the parsing speed on sentences no longer than 10.}
\label{tab:speed}
\end{table}
\subsection{Running Speed}

Our model have a time complexity of $O(d_u^2+d_b^2+n^3)$ while the first order model of \citet{dozat2018simpler} has a time complexity of $O(d_u^2+n^2)$ in scoring and decoding (where $d_u$ and $d_b$ are the hidden sizes of the biaffine layer and trilinear layer and $n$ is the sentence length). We compare these models with respect to training speed and parsing speed on an Nvidia Tesla P40 server. The result is shown in Table \ref{tab:speed}. Mean field variational inference slows down training and parsing by 35\% and 20\% respectively compared with the baseline. However, loopy belief propagation slows down training and parsing by 65\% and 67\% respectively compared with the baseline.

\subsection{Significance Test} We trained 25 basic models of our approach and the baseline with the same hyperparameters in Table \ref{tab:hyper_both} on each formalism. Student's t-test shows that our second-order model outperforms our baseline model on all the formalisms with a significance level of 0.005. 

\section{Related Work}
\subsection{Semantic Dependency Parsing}
Semantic dependency parsing can be classified as transition-based approaches and graph-based approaches. \citet{wang2018neural} proposed a transition-based parser for semantic parsing, while \citet{du2015peking} proposed a hybrid parser that benefits from both transition-based approaches and graph-based approaches. \citet{peng2017deep} proposed a graph-based approach that trains on all the three formalisms simultaneously and \citet{peng2018learning} further proposed to learn from different corpora. \citet{dozat2018simpler} proposed a graph-based simple but powerful neural network for semantic dependency parsing using a bilinear or biaffine \cite{dozat2016deep} layer to encode the interaction between words. Most of these approaches proposed first-order dependency parser while \citet{martins2014priberam} proposed a way to encode higher-order parts with hand-crafted features and introduced a novel co-parent part for semantic dependency parsing. They used discrete optimizing algorithm alternating directions dual decomposition ($\text{AD}^3$) as their decoder. \citet{cao2017quasi} also proposed a quasi-second-order semantic dependency parser with dynamic programming. 
Our model contains second-order information comparing with the first-order approaches and benefits from end-to-end training comparing with other second-order approaches.
\subsection{Higher-Order Dependency Parsing}
Higher-order parsing has been extensively studied in the literature of syntactic dependency parsing. Much of these work is based on the first-order maximum spanning tree (MST) parser of \citet{mcdonald2005online} which factorizes a dependency tree into individual edges and maximizes the summation of the scores of all the edges in a tree. \citet{mcdonald2006online} introduced a second-order MST that factorizes a dependency tree into not only edges but also second-order sibling parts, which allows interactions between adjacent sibling words. \citet{carreras2007experiments} defined second-order grandparent parts representing grandparental relations. \citet{koo2010efficient} introduced third-order grand-sibling and tri-sibling parts. A grand-sibling part represents a grandparent with two grandchildren and a tri-sibling part represents a parent with three children. \citet{ma2012fourth} defined grand-tri-sibling parts for fourth-order dependency parsing.

Many previous approaches to higher-order dependency parsing perform exact decoding based on dynamic programming, but there is also research in approximate higher-order parsing. \citet{martins2011dual} proposed an alternating directions dual decomposition ($\textrm{AD}^3$) algorithm which splits the original problem into several local sub-problems and solves them iteratively. They employed $\text{AD}^3$ for second-order dependency parsing to speed up decoding. \citet{smith2008dependency} and \citet{gormley2015approximation} proposed to use belief propagation for approximate higher-order parsing, which is closely related to our work.

While higher-order parsing has been shown to improve syntactic dependency parsing accuracy, it receives less attention in semantic dependency parsing.
\citet{martins2014priberam} proposed second-order semantic dependency parsing and employed $\textrm{AD}^3$ for approximate decoding.
\citet{cao2017quasi} proposed a quasi-second-order parser and used dynamic programming for decoding with time complexity of $O(n^4)$.

\subsection{CRF as Recurrent Neural Networks}
\citet{zheng2015conditional} are probably the first to propose the idea of unfolding iterative inference algorithms on CRFs as a stack of recurrent neural network layers. They unfolded mean field variational inference in a neural network designed for semantic segmentation. There is a lot of subsequent work that employs this technique, especially in the computer vision area. For example, \citet{zhu2017structured} proposed a structured attention neural model for Visual Question Answering with a CRF over image regions and unfolded both mean field variational inference and loopy belief propagation algorithms as recurrent layers.

\section{Conclusion}
We proposed a novel graph-based second-order parser for semantic dependency parsing. We constructed an end-to-end neural network that uses a trilinear function to score second-order parts and finds the highest-scoring parse graph by either mean field variational inference or loopy belief propagation algorithms unfolded as recurrent neural network layers. Our experimental results show that our model outperforms previous state-of-the-art model and has higher accuracies especially on out-of-domain data and long sentences. Our code is publicly available at \url{https://github.com/wangxinyu0922/Second_Order_SDP}
\section*{Acknowledgments}
This work was supported by the Major Program of Science and Technology Commission Shanghai Municipal (17JC1404102).
\bibliography{acl2019}
\bibliographystyle{acl_natbib}

\end{document}